\pdfoutput=1
\documentclass[11pt]{article}

\usepackage{ACL2023}

\usepackage{times}
\usepackage{latexsym}

\usepackage[T1]{fontenc}
\usepackage[utf8]{inputenc}

\usepackage{microtype}

\usepackage{inconsolata}

\usepackage{booktabs}

\usepackage[dvipsnames]{colortbl}
\definecolor{Gray}{gray}{0.9}

\usepackage{soul}
\usepackage{makecell}

\catcode`\"=13     % change the category code of "
\def"#1"{``#1''}   % and make "stuff" expand to ``stuff''

\usepackage{adjustbox}

\usepackage{natbib}
\usepackage{multibib}
\makeatletter
\def\@mb@citenamelist{cite,citep,citet,citealp,citealt,citepalias,citetalias}
\makeatother
\newcites{languageresource}{~}

\usepackage{graphicx}
\usepackage{tabularx}
\usepackage{soul}

\title{Astro-NER --- Astronomy Named Entity Recognition:\\ Is GPT a Good Domain Expert Annotator?}

\author{Julia Evans \and Sameer Sadruddin \and Jennifer D'Souza \\
 TIB - Leibniz Information Centre for Science and Technology \\ Welfengarten 1B, 30167 Hanover, Germany}
\begin{document}
\maketitle
\begin{abstract}
In this study, we address one of the challenges of developing NER models for scholarly domains, namely the scarcity of suitable labeled data.  We experiment with an approach using predictions from a fine-tuned LLM model to aid non-domain experts in annotating scientific entities within astronomy literature, with the goal of uncovering whether such a collaborative process can approximate domain expertise. Our results reveal moderate agreement between a domain expert and the LLM-assisted non-experts, as well as fair agreement between the domain expert and the LLM model's predictions. In an additional experiment, we compare the performance of finetuned and default LLMs on this task.  We have also introduced a specialized scientific entity annotation scheme for astronomy, validated by a domain expert. Our approach adopts a scholarly research contribution-centric perspective, focusing exclusively on scientific entities relevant to the research theme. The resultant dataset, containing 5,000 annotated astronomy article titles, is made publicly available.
\end{abstract}

\section{Introduction}

Named Entity Recognition (NER) is an essential tool in modern NLP pipelines, facilitating many downstream tasks.  One such application is extracting information for populating knowledge graphs (KG) and other digital information structures \citep{cs-ner}.  However, there is a persistent bottleneck limiting the development of KGs for scientific disciplines: scholarly-focused NER poses unique challenges not addressed by generic NER solutions \citep{cs-ner, geo-ner}.

For general purpose NER, there is an abundance of labeled English text data, along with readily accessible NER tools. However, in the context of highly specialized scholarly domains, even English may present a low-resource data scenario when appropriately labeled data is rare.  Technical jargon and the particular stylistics of academic writing, as well as unique entities-of-interest beyond those common in NER, render existing NER datasets and tools difficult to apply \citep{geo-ner}.  At the same time, generating high-quality labeled data in scholarly domains is especially challenging due to a limited pool of qualified annotators
%--who are also well versed in the field of computational linguistics--
and their potential reticence to participate in the annotation process.  

The surge in the development of Large Language Models (LLMs) in recent years has led to research investigating whether LLMs can support, or even replace, annotators, thus easing the burden of generating labeled text data \citep{wang-etal-2021-want-reduce, ding-etal-2023-gpt}.  Such experiments are particularly relevant for annotation tasks requiring uncommon or highly-specialized expert knowledge.  

In this work, we present an approach to address the bottleneck issue of limited expert annotator availability.  We use predictions from a finetuned GPT-3.5 model \citep{gpt3} to support non-domain expert annotators in the task of annotating scientific entities in astronomy literature.  This is a highly complex task for non-domain experts.  Scholarly papers in astronomy contain a large set of scientific terminology for celestial objects, astronomical phenomena, and astrophysical concepts, which must be understood with precision and nuance.  Moreover, these concepts must be interpreted within the broader context of astronomy research, which requires knowledge of the objectives of a research work, the significance of its findings, and its connection with previous works and established theories.  Given these complexities, on the one hand, domain expert annotators are preferred. On the other hand, however, an approach which allowed non-domain expert annotators to perform at a similar level of competency would greatly benefit the speed at which such specialized datasets could be produced.

The contributions of this paper are as follows.  1) The release of a corpus of titles from astronomy literature annotated with the scientific entities that reflect the contribution of the work. 2) A quantitative empirical assessment of three LLM variants in both default and finetuned states, for our defined astronomy scientific named entity recognition task. Additionally, state-of-the-art reported models for a different NER task for astronomy were also finetuned to our task dataset and released as baseline results.  3) An evaluation of the feasibility of using LLMs to assist in annotation tasks where specialized knowledge is required but no domain expert is available.

\section{Related Work} \label{sec:related-work}

There are multiple data labeling and structuring approaches that can be followed when creating scholarly knowledge representations.  \citet{grezes-etal-2022-overview} organized the DEAL shared task focused on astronomy NER in which they propose a set of thirty-three entities and enlist a domain expert to annotate text fragments from scholarly articles in astronomy.  They divide their labels into five categories: 1) generic NER entities (Person,\footnote{Throughout this paper, we use UpperCamelCase to indicate entity labels.} Organization, Location); 2) entities related to facilities for studying astrophysics (Observatory, Telescope); 3) entities related to funding (Grant, Proposal); 4) astronomical object entities (CelestialObject); and 5) entities found in academic literature (Citation, URL).

We take a somewhat different perspective and follow previous work in \textit{contribution-centric NER} \citep{cs-ner, agri-ner} by focusing exclusively on the entities pertinent to describing the research contribution of a paper.  In this approach, ``only those entities that are either the outcome of a particular research endeavor or used to support the outcome of that work are candidate extraction targets'' \citep{cs-ner}. To that end, an appropriate annotation scheme encompasses both domain-specific entities such as AstrObject and PhysicalQuantity, as well as more domain-agnostic research-focused entities like ResearchProblem and Method.

However, annotation of this nature is often an expensive and time-consuming process, made all the more difficult by the challenge of finding, recruiting, and financing annotators who possess the required specialized knowledge. \citet{wang-etal-2021-want-reduce} have proposed an approach that potentially offers a solution to this predicament, demonstrated in their experiments involving GPT-3 generated labels for various NLP datasets. Their findings indicate that optimal performance is attained through a combination of GPT-3 and human annotators, specifically by establishing a minimum confidence threshold for GPT-3 predictions and subsequently undertaking manual relabeling for instances falling below that threshold.

In another study, \citet{ding-etal-2023-gpt} explored three approaches utilizing GPT-3 for annotating or generating training data for NER and other NLP tasks. Their research demonstrates that an NER model trained on a moderate volume of GPT-3 generated data (at least 1500 samples) outperforms a model trained on a comparatively smaller dataset of human-annotated data (100 samples), which is what might be obtainable for a similar cost and time expenditure. The best performance was achieved by leveraging \href{https://www.wikidata.org/wiki/Wikidata:Main_Page}{Wikidata} to extract example entities and prompting GPT-3 to generate sentences using said entities; the worst performance was using GPT-3 purely for annotating existing unlabeled data.

Finally, \citet{hedderich-etal-2021-survey} note that domain-adaptation by finetuning a general-domain model is a common solution that improves performance on tasks within the target domain.  Given that scholarly domains are highly specialized, such domain-adaptation may be relevant to the task of LLM annotation as well.

\section{Our Corpus}

\subsection{Task Definition}

In this paper, we present our Astro-NER corpus: a collection of 5000 astronomy article titles annotated with contribution-centric scientific entities.  This corpus was constructed, in part, by finetuning a GPT-3.5 model (see Section \ref{sec:anno-asst}) for the task of astronomy literature annotation, and then making the predictions available to non-domain expert annotators.  Rather than using a confidence threshold for determining when to accept the GPT labels (as in \citet{wang-etal-2021-want-reduce}), annotators considered every label and used their own judgement.  The labels and their definitions can be found in Table \ref{tab:labels} (see Section \ref{sec:anno-process} for more information on how these labels were selected).

Our data source consists of the titles from around 15,000 astronomy articles with the CC-BY redistributable license, downloaded from Elsevier.  From this, approximately 5000 titles were randomly selected for annotation by two graduate students.  Annotator 1 is a PhD student in Computational Linguistics and Annotator 2 is a master's student in Computer Science.  Both annotators possess advanced proficiency in scientific English.

\begin{table*}[t]
\centering
\begin{tabular}{p{0.2\linewidth} p{0.75\linewidth}}
\toprule
\textbf{Label} & \textbf{Definition}\\
\midrule
AstrObject & All concepts representing astronomical objects, e.g. black holes.\\
\rowcolor{Gray}
AstroPortion & All concepts representing portions of astronomical objects which are not astronomical objects themselves, e.g. sunspots.\\
ChemicalSpecies & Atomic elements such as element names from the periodic table, atoms, nuclei, dark matter, e.g. Fe.\\
\rowcolor{Gray}
Instrument & Names of measurement instruments, including  telescopes, e.g. Large Hadron Collider.\\
Measurement & Measured observational parameters or properties (both property and value), e.g. frequency.\\
\rowcolor{Gray}
Method & Abstractions which are commonly used to support the solution of the investigation, e.g. minimal supersymmetrical model.\\
Morphology & Geometry or morphology of astronomical objects or physical phenomena, e.g. asymmetrical.\\
\rowcolor{Gray}
PhysicalQuantity & Properties of physical phenomena interacting, e.g. gravity.\\
Process & Phenomenon or associated process, e.g. Higgs boson decay.\\
\rowcolor{Gray}
Project & Survey or research mission, e.g. the dark energy survey.\\
ResearchProblem & The theme of the investigation, e.g. final state hadronic interactions.\\
\rowcolor{Gray}
SpectralRegime & Observed or analyzed electromagnetic spectrum, e.g. mega electron volt.
\\
\bottomrule
\end{tabular}
\caption{The final scientific entity labels applicable on titles of scholarly articles in the astronomy domain and their definitions. In case of overlap, ResearchProblem is selected over all other entity types, and Method is selected over all other entity types except ResearchProblem.} 
\label{tab:labels}
\end{table*}

\subsection{Annotation Process} \label{sec:anno-process}

Both the scientific entity labels and the annotation guidelines were refined through an iterative process. An initial list of candidate labels was drawn from the definitions in previous works on astronomy NER \citep{becker-et-al, murphy-etal-2006-named, grezes-etal-2022-overview}, contribution-centric NER \citep{cs-ner, agri-ner}, and the top-level concepts in an astronomy-specific ontology \citep{ivoa}.  This yielded 36 potential labels, of which not all were suited to our task.   Some labels from previous contribution-centric NER do not apply to the domain of astronomy, such as Language and Dataset.  Additionally, several labels could be subsumed under broader categories, such as Star, Planet, and Nebula under the label of AstrObject.  After removing such labels which were too fine-grained for our purposes or unrelated to our task, a set of 21 labels remained.

\begin{table*}[t]
\centering
\begin{tabular}{p{0.3\linewidth}p{0.3\linewidth}p{0.3\linewidth}}
\toprule
\textbf{Round I} & \textbf{Round II} & \textbf{Round III}\\
\midrule
AstrObject, AstroPortion, Atomic Element, \st{Classification Category}, \st{Dataset}, Date, Duration, EMS Spectrum Range, Frequency, \st{Galaxy}, Instrument Name, \st{Ion}, \st{Language}, Location, Luminosity, Measurement, Method, Morphology, \st{Nebula}, \st{Planet}, Position, \st{Process-1}, Process-2, Research Problem, \st{Resource}, \st{Solution}, Source Name, Source Type, Spectral Feature, \st{Star}, \st{Star Cluster}, \st{Supernova}, Survey, \st{Technology}, Telescope, \st{Tool} & \cellcolor{Gray} AstrObject, AstroPortion, Atomic Element, \st{Date}, \st{Duration}, \st{EMS Spectrum Range}, \st{Frequency}, Instrument Name, \st{Location}, \st{Luminosity}, Measurement, Method, Morphology, \st{Position}, Process, Research Problem, \st{Source Name}, \st{Source Type}, \st{Spectral Feature}, \st{Survey}, \st{Telescope} \newline +Force, +Matter, +Model & AstrObject, AstroPortion, \st{Atomic Element} → Chemical Species, \st{Force} → Physical Quantity, Instrument, \st{Matter}, Measurement, Method, \st{Model}, Morphology, Process, Research Problem \newline
+Project, +Spectral Regime\\
\bottomrule
\end{tabular}
\caption{The evolution of the label set after each round of discussions.  Round I shows the entire list of candidate labels, with those deemed too fine-grained or irrelevant for our task crossed out.  Round II shows the label set after discussing the first pilot annotation, in which labels were removed if unused and additional labels added.  Round III shows the final label set after discussing the second pilot annotation with a domain expert.} 
\label{tab:label-selection}
\end{table*}

A small pilot annotation was performed on a random sample of 50 titles to evaluate the coverage and applicability of the labels.  After examining the results, it was decided to tentatively remove 12 unused labels, and add an additional 3 labels for relevant entities which were not covered under the existing definitions, resulting in 12 labels.  The unused labels were primarily concerned with specific measurements or data properties such as Duration, Luminosity, and Spectral Feature. Subsequently, a second pilot annotation was performed on a random sample of 100 titles with the new set of labels.

For the final assessment, the astronomy subject librarian at the German National Library of Science and Technology was consulted on the conceptual relevance and technical validity of the selected labels for the field of astronomy.  On his advice, 2 labels were renamed, 2 were removed, and 2 new labels were added.  Additionally, feedback on the 100 annotated titles provided explicit training in accurately selecting entities.  See Table \ref{tab:label-selection} for an overview of the selection process.

Annotation guidelines were also developed in tandem with the labels.  As a foundation, we used the annotation guidelines from previous work in contribution-centric NER \citep{cs-ner}, which defined the linguistic and contextual considerations for identifying entity types and spans.  After each round of pilot annotations, the guidelines were reviewed and some small adaptations made.

One particularity of our annotation scheme is that some entities may correspond to both ResearchProblem or Method and another label.  For instance, "jet quenching" is a phenomenon in which some types of particles produced in the early stages of a collision lose energy as they traverse the collision-created medium.  This is a Process according to our label definitions.  In the case that it is the subject of the investigation, it would also be a ResearchProblem.  As only one label per entity is allowed, we decided on a precedence hierarchy in which ResearchProblem is selected above any other labels which may apply, and Method is selected above any other labels except ResearchProblem.  These two entity types are prioritized given their centrality to contribution-centric knowledge representations.

The most significant principles guiding our annotation process can be summarized as follows.

\begin{enumerate}
    \item There are no restrictions on the morphosyntactic form of entities, but noun phrases without articles are preferred wherever possible.
    \item Include prepositions only if they are indeed part of the term itself or modify the entity in an essential way.
    \item Select the most precise text reference possible, including all necessary modifiers, as a single unit.  Consider the intended meaning in the given context to determine whether a modifier is necessary -- anything that changes the meaning of a term ought to be included.
    \item Given an expression in which several concepts or terms are nested or containing conjunctions with ellipsis of a shared noun phrase, annotate the entire sequence as one entity.
    \item Follow the precedence hierarchy of \textit{ResearchProblem} > \textit{Method} > all other entity types.
\end{enumerate}

After finalizing the labels and guidelines, the annotation task was conducted in two phases.

\textbf{Phase I.} Annotators determined entities by reading the paper's abstract, looking up definitions of terms, and/or consulting ChatGPT.

\textbf{Phase II.} Annotators were provided with predicted labels from a finetuned GPT-3.5 model for each title.  After checking the predictions, annotators could use any of the strategies from Phase I as well, and were free to accept or disregard any of the predictions.

The final corpus contains 5000 annotated texts.  Table \ref{tab:corpus-size} summarizes the distribution across annotators and annotation settings, and Table \ref{tab:entity-types} shows the frequency of entity types.

\begin{table}
\centering
\begin{tabular}{lcc|c}
\toprule
& \textbf{Phase I} & \textbf{Phase II} & \textbf{Total}\\
\midrule
\textit{Annotator 1} & 2325 & 1583 & 3908 \\
\textit{Annotator 2} & 98 & 994 & 1092 \\
\midrule
\textit{Total} & 2423 & 2577 & 5000\\
\bottomrule
\end{tabular}
\caption{The size of our corpus.} 
\label{tab:corpus-size}
\end{table}

\begin{table*}[t]
\centering
\begin{tabular}{cccccccccccc}
\toprule
\stackbox[c]{\scriptsize \textbf{Astro\\ Object}} & \stackbox[c]{\scriptsize \textbf{Astro\\ Portion}} & \stackbox[c]{\scriptsize \textbf{Chem.\\ Species}} & \stackbox[c]{\scriptsize \textbf{Inst-\\ rument}} & \stackbox[c]{\scriptsize \textbf{Meas-\\ urment}} & \stackbox[c]{\scriptsize \textbf{Method}} & \stackbox[c]{\scriptsize \textbf{Morph-\\ ology}} & \stackbox[c]{\scriptsize \textbf{Phys.\\ Quant.}} & \stackbox[c]{\scriptsize \textbf{Process}} & \stackbox[c]{\scriptsize \textbf{Project}} & \stackbox[c]{\scriptsize \textbf{Research\\ Problem}} & \stackbox[c]{\scriptsize \textbf{Spect.\\ Regime}}\\
\midrule
143 & 97 & 851 & 615 & 320 & 3169 & 385 & 547 & 1273 & 123 & 3801 & 141\\
\bottomrule
\end{tabular}
\caption{The frequency of occurrences of scientific entity types in our corpus.} 
\label{tab:entity-types}
\end{table*}

\subsection{Finetuned GPT-3.5 as an Annotation Assistant} \label{sec:anno-asst}

In Phase I, an initial 2001 texts were annotated by a single annotator and used to finetune the GPT-3.5 model \texttt{davinci-002}\footnote{https://platform.openai.com/docs/models} to predict our astronomy labels.  A two-stage finetuning process was used: 

\begin{enumerate}
    \item A prompt containing an explanation of the task, all entity types and their definitions, and a few rules for annotation such as no overlapping spans was used to finetune the model on 100 texts.
    \item A much shorter prompt containing a single sentence of task instruction and the list of entity types without definitions was used for a second round of finetuning the previous model on 1901 texts.
\end{enumerate}

\begin{table*}[t]
\centering
\begin{tabular}{l|p{0.42\linewidth} p{0.42\linewidth}}
\toprule
& \textbf{Stage I} & \textbf{Stage II}\\
\midrule
\textit{Intro} & Please fulfill the following NER task by annotating the given scholarly paper title in the domain of astronomy. ... & Please fulfill the following NER task by annotating the given scholarly paper title in the domain of astronomy.\\
\rowcolor{Gray}
\textit{Entities} & Entity types to consider: 1. AstrObject: subsumes all the concepts representing astronomical objects. ... & Consider only the following 12 entity types and rely on your knowledge for their definitions: 1. AstrObject, 2. AstroPortion, ...\\
\textit{Rules} & Annotation rules: - Each word can be included in at most one annotation. ... & Rely on your knowledge of the annotation rules\\
\rowcolor{Gray}
\textit{Output} & Please provide the annotations in JSON format with the entity labels as keys. Annotate the following title: "{title}" & please provide the annotations in JSON format with the entity labels as keys. Annotate the following title: "{title}"\\
\bottomrule
\end{tabular}
\caption{Skeleton of the prompts used for finetuning GPT-3.5.} 
\label{tab:prompts}
\end{table*}

The resulting finetuned GPT-3.5 model was used in Phase II of the annotation process to predict labels for an additional 2577 texts.  See Table \ref{tab:prompts} for a skeleton outline of the prompts.

\subsection{Qualitative Observations}

\paragraph{Task Difficulty.}  Several features of astronomy literature make the annotation task particularly difficult for non-experts.  Below is a summary of some of the challenges.

\begin{itemize}
    \item Lists of concepts or phenomena without any explicit relationship between them, e.g. "Generalized Poincaré algebras and Lovelock–Cartan gravity theory".
    \item The form “$\langle$method/process$\rangle$ $\langle$connector$\rangle$ $\langle$method/process$\rangle$” where it is unclear whether a method is being applied in a certain context to understand or develop the method itself or whether it is being used to learn more about the process, e.g. "One-loop QCD corrections to the $e + e - \rightarrow W + W - b b ^{-}$ process".
    \item The research problem is implied but not explicitly stated, e.g. "Quasi-normal modes of holographic system with Weyl correction and momentum dissipation" (Quasi-normal modes are a concept for studying black holes and strongly coupled systems).
    \item Metonymy in which the actual term and the intended referent correspond to different labels, e.g. "Complementarity between Hyperkamiokande and DUNE in determining neutrino oscillation parameters" (Hyperkamiokande and DUNE are both instruments, but the implied meaning is "measurements from Hyperkamiokande/DUNE").
    \item The linguistic structure obscures the roles, e.g. "Transverse anomalies and Dyson–Schwinger equation in QED3 and QED2 theories" (the Dyson–Schwinger equation is used to study transverse anomalies in the framework of QED3 and QED2 theories).
\end{itemize}

For domain experts, the research applications of different methods and the relationships between them are likely clear, regardless of how they are formulated in the text.  But for non-experts, a considerable amount of deciphering may be required.

\paragraph{Finetuned GPT-3.5 Performance.}

Table \ref{tabel:gpt-anno-analysis} shows examples of some of the most common types of errors made by the finetuned GPT-3.5 model.  Occasional errors also include reordering words and creating new labels (a proposed theory called "Gravity's Rainbow" was labeled as \textit{Book}).  The predictions are generally highly plausible, even in cases where they are not totally correct.

The two annotators had different perceptions of the utility of the finetuned GPT-3.5 predictions. Annotator 1 found them helpful for narrowing down the potential entities and labels, while then using her own judgement to refine the final annotations.  Meanwhile, Annotator 2 agreed the predictions were a good starting reference, but found some of the errors to be distracting and the overall predictions not trustworthy.

\begin{table*}[!htb]
\centering
\begin{tabular}{p{0.15\textwidth}p{0.4\textwidth}p{0.4\textwidth}}
\toprule
 & \bf Predictions & \bf Annotations \\ \midrule

\textit{Incorrect} & Effective theory of {\textcolor{teal}{dark matter decay}} into {\textcolor{violet}{monochromatic photons}} and its implications: Constraints from associated cosmic-ray emission. \newline {\color{teal}{ResearchProblem}}, {\color{violet}{Method}} & {\textcolor{violet}{Effective theory of dark matter decay into monochromatic photons}} and its implications: Constraints from associated {\textcolor{teal}{cosmic-ray emission}}. \newline {\color{violet}{Method}}, {\color{teal}{ResearchProblem}}\\

\rowcolor{Gray}
\textit{Under labeled} & Energy conditions in F ( T , {\color{orange}{$\Theta$}} ) gravity and compatibility with a stable de Sitter solution. \newline {\color{orange}{PhysicalQuantity}} & {\textcolor{teal}{Energy conditions in F ( T , $\Theta$ ) gravity}} and compatibility with a {\textcolor{violet}{stable de Sitter solution}}. \newline {\color{teal}{ResearchProblem}}, {\color{violet}{Method}}\\

\textit{Under specified} & The origin of {\textcolor{teal}{large-p T $\pi$ $\theta$ suppression}} at {\textcolor{purple}{RHIC}}. \newline  {\color{teal}{ResearchProblem}}, {\color{purple}{Instrument}} & The {\textcolor{teal}{origin of large-p T $\pi$ $\theta$ suppression}} at {\textcolor{purple}{RHIC}}. \newline {\color{teal}{ResearchProblem}}, {\color{purple}{Instrument}}\\

\rowcolor{Gray}
\textit{Over specified} & {\textcolor{teal}{Lepton flavor violation in the triplet Higgs model}}. \newline  {\textcolor{teal}{ResearchProblem}} & {\textcolor{teal}{Lepton flavor violation}} in the {\textcolor{violet}{triplet Higgs model}}. \newline {\color{teal}{ResearchProblem}}, {\color{violet}{Method}}\\

\textit{Missing} \newline \textit{coordinated} \newline \textit{expressions} & Baryon number and {\textcolor{teal}{lepton universality violation}} in leptoquark and {\textcolor{violet}{diquark models}}. \newline {\color{teal}{ResearchProblem}}, {\color{violet}{Method}}  & {\textcolor{teal}{Baryon number and lepton universality violation}} in {\textcolor{violet}{leptoquark and diquark models}}. \newline {\color{teal}{ResearchProblem}}, {\color{violet}{Method}}\\
\bottomrule
\end{tabular}
\caption{Common prediction error types made by finetuned GPT-3.5.}
\label{tabel:gpt-anno-analysis}
\end{table*}

\subsection{Inter-Annotator Agreement}

Inter-annotator agreement was computed on two sets of 100 texts using Cohen's $\kappa$, with all tokens included.  The first set of texts were annotated during Phase I, while the second set were annotated during Phase II.  The domain expert was only available to annotate one set of texts, for which the set from Phase II was chosen.  However, he did not have access to the finetuned GPT-3.5 predictions and rather followed the annotation procedure from Phase I.  The results are shown in Table \ref{tab:iaa}.

\begin{table}
\centering
\begin{tabular}{lccc}
\toprule
& \textbf{A1-A2} & \textbf{A1-DE} & \textbf{A2-DE}\\
\midrule
\textit{Phase I} & 0.62 & - & - \\
\textit{Phase II} & 0.53 & 0.42 & 0.35 \\
\bottomrule
\end{tabular}
\caption{Cohen's $\kappa$ for both annotators (A1 and A2) and the domain expert (DE), computed on 100 texts.  The domain expert did not have access to the finetuned GPT-3.5 predictions during the annotation process.} 
\label{tab:iaa}
\end{table}

The scores between the two annotators indicate moderate agreement, reflecting the difficulty and complexity of this task.  Of note is the finding that agreement decreased between Phase I and Phase II, indicating that the finetuned GPT-3.5 predictions biased the annotators towards different conclusions.

Meanwhile, the scores between the domain expert and Annotator 1 have low moderate agreement, whereas the domain expert and Annotator 2 have fair agreement.  These results indicate that even with support from the finetuned GPT-3.5 model, non-domain experts can still only weakly approximate the performance of a domain expert.

\begin{table}
\centering
\begin{tabular}{lcc}
\toprule
& \textbf{GPT-3.5 OOTB} & \textbf{GPT-3.5 FT}\\
\midrule
\textit{A1} & 0.12 & 0.70\\
\textit{A2} & 0.10 & 0.48\\
\textit{DE} & 0.14 & 0.35\\
\bottomrule
\end{tabular}
\caption{Cohen's $\kappa$ computed between annotators or the domain expert and GPT-3.5 out-of-the-box or GPT-3.5 finetuned.}
\label{tab:iaa-gpt}
\end{table}

Table \ref{tab:iaa-gpt} shows the agreement between the human annotators and the GPT-3.5 models.\footnote{Due to limitations in the GPT data (see Section \ref{sec:exp}), 57 texts were included in the calculations for GPT-3.5 OOTB and 98 for GPT-3.5 FT.}  We observe very low agreement between all annotators and GPT-3.5 out-of-the-box, whereas agreement with the finetuned model varies significantly.  Annotator 1 showed substantial agreement and Annotator 2 showed moderate agreement with the finetuned model, which aligns with their reported experiences of working with the predictions.  Meanwhile, the domain expert and finetuned model have an agreement of 0.35, indicating only fair agreement.  This provides additional evidence that even with finetuning, GPT-3.5 still lacks the domain knowledge and sophistication to perform annotation at a level comparable to a domain expert.

\section{Experiments and Results}\label{sec:exp}

\paragraph{Dataset.} An experimental dataset of 1500 texts was used to compare the performance of out-of-the-box GPT-3.5 (GPT-3.5 OOTB), out-of-the-box GPT-4 (GPT-4 OOTB), and finetuned GPT-3.5 (GPT 3.5 FT).  All texts came from Phase II and were divided evenly between the annotators.  Predictions from each model were obtained using the same prompt as in stage 2 of the finetuning process (see Section \ref{sec:anno-asst}).

For some texts, GPT failed to find any entities and these texts are therefore excluded.  In other cases, entities with overlapping spans were returned.  Here we used a precedence hierarchy similar to that of our human annotators to manually resolve the labels: 1) any spans overlapping with ResearchProblem are discarded; 2) any spans except ResearchProblem overlapping with Method are discarded; 3) for all other overlapping spans, the first predicted label is taken and the rest discarded.  Finally, some texts required additional processing to be made usable, and are also excluded.  

There was a stark difference in quality between GPT3.5 OOTB and the other two models.  The response object very often contained malformed json, frequently so mangled it was impossible to process.  Additionally, it had a tendency to return all labels in the same order they were passed in the prompt, with an annotation for each one.  This made our precedence hierarchy impractical, since the output order inherently privileged certain labels over others.  As a result of these constraints, there are significantly fewer usable texts from the GPT3.5 OOTB model.

\paragraph{GPT Models.}
The usable texts from each model were aligned with the corresponding human-annotated texts so that the predictions could be compared against our corpus.  This resulted in 793 usable texts for GPT-3.5 OOTB, 1497 for GPT3.5 FT, and 1465 GPT-4 OOTB.  Micro averages for precision, recall, and f-score are reported in Table \ref{tab:p-r-f}.  

The results indicate extremely weak performance by GPT-3.5 OOTB.  GPT-4 OOTB shows an impressive 19-point improvement, while the finetuned model performs by far the best. Nonetheless, for an NER-adjacent task, an f-score of 0.51 may be considered low.

\begin{table}
\centering
\begin{tabular}{lccc}
\toprule
& \textbf{P} & \textbf{R} & \textbf{F1}\\
\midrule
\textit{GPT-3.5 OOTB} & 0.04 & 0.04 & 0.04 \\
\rowcolor{Gray}
\textit{GPT-3.5 FT} & \textbf{0.55} & \textbf{0.48} & \textbf{0.51} \\
\textit{GPT-4 OOTB}  & 0.23 & 0.24 & 0.23 \\
\bottomrule
\end{tabular}
\caption{Micro precision (P), recall (R), and F1-score (F1) evaluating GPT predictions against human annotations. Annotators saw the GPT-3.5 FT predictions during the annotation process.} 
\label{tab:p-r-f}
\end{table}

\paragraph{NER Models.}

As the ultimate goal of this annotation project is to provide training data for an Astro-NER service, the following NER models were also trained and evaluated.  The FLAN-T5 model \citep{flan-t5} in the Small (77M) size was selected due to its efficiency at learning new tasks \citep{pmlr-v202-longpre23a}.  Additionally, the mT5 model \citep{mt5} in the Small (300M) size was also included because the best performing system \citep{ghosh-etal-2022-astro} from the DEAL astronomy NER shared task \citep{grezes-etal-2022-overview} (described in Section \ref{sec:related-work}) utilized this model.  The same hyperparameters were used across models: 100 epochs, learning rate of 3e-4, and a batch size of 16.

The micro precision, recall, and f-score metrics for each of our NER models along different dataset splits are presented in Table \ref{tab:ner-p-r-f}.  Overall, the best results are obtained with mT5 and a random 90/10 split of the complete dataset for training and testing, with an f-score of 0.43.  For reference, the top performing system on the DEAL task reported an f-score of 0.81, although it must be noted that this task used a different dataset with a different annotation scheme, so the results are not directly comparable.  Nonetheless, we conclude that our results are not competitive in the context of current astronomy NER systems.  We hypothesize that having two annotators lacking expertise in the domain may have introduced some inconsistencies into the dataset which were reflected in the training results of the model.

\begin{table}
\centering
\begin{tabular}{lccc}
\toprule
& \textbf{P} & \textbf{R} & \textbf{F1}\\
\midrule
\textit{mT5-Small$_{1}$} & 0.38 & 0.32 & 0.35 \\
\rowcolor{Gray}
\textit{FLAN-T5-Small$_{1}$} & 0.36 & 0.33 & 0.34 \\
\midrule
\textit{mT5-Small$_{2}$} & 0.41 & \textbf{0.43} & 0.42 \\
\rowcolor{Gray}
\textit{FLAN-T5-Small$_{2}$} & 0.37 & 0.39 & 0.38 \\
\midrule
\textit{mT5-Small$_{3}$} & \textbf{0.45} & 0.42 & \textbf{0.43} \\
\rowcolor{Gray}
\textit{FLAN-T5-Small$_{3}$} & 0.40 & 0.41 & 0.40 \\
\bottomrule
\end{tabular}
\caption{Micro precision (P), recall (R), and F1-score (F1) for each of our NER models.  The subscript numbers indicate the dataset split: $_{1}$ trained on the same 2000 texts as the GPT-3.5 FT model and tested on the same 1500 experimental texts; $_{2}$ trained on all texts except the 1500 experimental texts and tested on those; $_{3}$ trained and tested on a random 90/10 split of the complete datatset.} 
\label{tab:ner-p-r-f}
\end{table}

\section{Discussion and Limitations}

Based on the precision, recall, and f-score metrics, we conclude the following.  GPT-3.5 OOTB is not a good domain expert annotator, which aligns with our intuition that it excels at handling common sense tasks but not tasks requiring domain expertise.  GPT-4 OOTB shows more promise, but is still insufficiently informed in highly-specialized scientific fields.  In order to use GPT as an annotation assistant, finetuning is necessary.  We find an enormous 47 point improvement in f-score before and after finetuning.  We also find that the finetuned GPT-3.5 outperforms our best NER model.  Nonetheless, the results overall are weak, and our best NER model underperforms compared to previous work in astronomy NER \citep{grezes-etal-2022-overview, ghosh-etal-2022-astro}.  Moreover, these f-scores are computed against the annotations of non-domain experts, whose annotations are themselves subject to validation.

Considering the inter-annotator agreement, we conclude that specialized scientific domains remain an area in which domain expert annotators are still necessary.  Annotator 1, whose annotations were slightly more aligned with the domain expert, benefited the most from the GPT assistance.  On the other hand, Annotator 2 seemed to maintain some independence from the GPT predictions and had slightly lower agreement with the domain expert as well.  But compared to the significant difference in agreement between the annotators and GPT (Cohen's $\kappa$ 0.70 vs 0.48), the difference between their agreement with the domain expert is relatively minor (Cohen's $\kappa$ 0.42 vs 0.35)--it seems that adherence to the GPT predictions had minimal impact on the accuracy of annotations for non-domain experts.

Overall, the agreement between this domain expert and the annotators may be considered low, despite the complexity of this task.  However, we also note that scientific entity annotation is an inherently subjective task.  For domains entailing high-expertise, allowance must be made for subjectivity in the annotation decisions, and we recognize that results with a different domain expert might differ.

We do observe one benefit to using GPT-3.5 as an annotation assistant: it dramatically quickened the pace of annotation.  Phase II of the annotation process was completed in just six weeks, whereas Phase I took approximately 4 months, despite a similar weekly time investment.  In this way, GPT can be thought of as a sounding board for annotators, giving them a starting point for consideration rather than a blank slate.  Nevertheless, this approach is only advantageous insofar as high-quality annotations can be obtained.

Our methodology was limited by the scant availability of the domain expert, which we note as a realistic setting for such projects.  As a result, our model was finetuned on non-domain expert annotations.  Expert-labeled training data might have resulted in a different outcome, but is not feasible in all annotation projects.

Some additional limitations concerning the original content of the dataset warrant discussion.  The domain expert noted that the titles were overwhelmingly from the astronomy subfield of astrophysics, with a particular emphasis on astroparticle physics.  There was discussion as to whether describing this as an astronomy dataset was inappropriately general, but given that the source of the titles was Elsevier publications labeled as astronomy, we chose to maintain this nomenclature.

The distribution of entity types is extremely unbalanced in our corpus.  Given our precedence hierarchy, as well as the conventions of academic title writing, ResearchProblem and Method appearing 3801 and 3169 times respectively is not unexpected.  However, only one other label appears more than 1000 times: Process, with 1273 instances.  The remaining entity types are mostly supported by several hundred samples.  We note that this significant disparity is not ideal.

Finally, the costs of the various models must be considered.  Getting predictions for the 1500 texts in the experimental dataset cost \$8.35 for GPT-3.5 OOTB and \$10.98 for GPT-4 OOTB.  The finetuned model was considerably more expensive, costing \$49.80 to finetune and \$33.63 to get predictions on the experimental texts (\$57.68 for all texts in Phase II), for a total of \$83.43 (or \$107.48 when including all texts).

\section{Conclusions}

In this work, we address one of the challenges associated with acquiring NER models for scholarly domains, namely the scarcity of appropriate labeled data.  While the involvement of domain experts in annotation projects is often indispensable due to the requisite subject knowledge, the reality is that access to such experts may be limited.  We present a novel approach to overcoming this hurdle by enlisting a finetuned GPT-3.5 model to assist non-domain experts in annotating scientific entities in astronomy literature.  On a small sample of the data, we find that the agreement between the domain expert and GPT-assisted non-experts is fair to moderate, while the agreement between the domain expert and the finetuned predictions is also fair.

As part of this endeavour, we have developed a scientific entity annotation scheme for astronomy and validated it with a domain expert.  Unlike previous works in astronomy NER, we take a contribution-centric perspective to scientific entity identification: we select only those entities which are pertinent to the theme of the investigation.  The dataset resulting from this annotation scheme, consisting of 5000 annotated titles from astronomy articles, is also published to support the continued development of scholarly contribution-focused astronomy NLP tools.

\section*{Ethics Statement}

%Scientific work published at EMNLP 2023 must comply with the \href{https://www.aclweb.org/portal/content/acl-code-ethics}{ACL Ethics Policy}. We encourage all authors to include an explicit ethics statement on the broader impact of the work, or other ethical considerations after the conclusion but before the references. The ethics statement will not count toward the page limit (8 pages for long, 4 pages for short papers).

In this work we have presented our Astro-NER corpus.  During its creation, we used a finetuned LLM.  
%~\cite{instructgpt,flan} is a research approach that aims to align language models with more desirable objectives and human preferences. Its purpose is to mitigate the issues associated with language models, such as toxic behavior~\cite{toxic1,toxic2,toxic3}, generation of non-factual information~\cite{unfact1,unfact2,unfact3}, and challenges in deployment and evaluation~\cite{fake1,fake2,fake3}.
In this context, we declare the instructions selected for finetuning in this study were intended to align the behavior of the language models towards producing responses that are both helpful (fulfilling our objective) and harmless (not causing any physical, psychological, or social harm to individuals or the environment). 

There were no living subjects analyzed in this study. Overall, this study complies with the \href{https://www.aclweb.org/portal/content/acl-code-ethics}{ACL Ethics Policy}.

\section*{Data and Code Availability}

To facilitate further research, our Astro-NER dataset is publicly released at the following \href{https://anonymous.4open.science/r/astro-ner-B064/datasets/astronomy_scientific_entity_dataset.json}{repository}, along with our experimental datasets. Furthermore, the prompts used to finetune GPT-3.5 are accessible \href{https://anonymous.4open.science/r/astro-ner-B064/prompts/stage1.txt}{here} and \href{https://anonymous.4open.science/r/astro-ner-B064/prompts/stage2.txt}{here}. The code used to finetune the mT5 and Flan-T5 models can be downloaded \href{https://anonymous.4open.science/r/astro-ner-B064/NER_models.ipynb}{here}.  The annotation guidelines can be viewed \href{https://anonymous.4open.science/r/astro-ner-B064/annotation_guidelines.pdf}{here}.

\section*{Acknowledgements}

This work was supported by the German BMBF project SCINEXT (ID 01lS22070).

\newpage
\section*{Bibliographical References}\label{sec:reference}

\bibliography{anthology,custom}

\begin{thebibliography}{15}
\expandafter\ifx\csname natexlab\endcsname\relax\def\natexlab#1{#1}\fi

\bibitem[{Becker et~al.(2005)Becker, Hachey, Alex, and Grover}]{becker-et-al}
Markus Becker, Benjamin Hachey, Beatrice Alex, and Claire Grover. 2005.
\newblock {Optimising Selective Sampling for Bootstrapping Named Entity Recognition}.
\newblock In \emph{Proceedings of the ICML-2005 Workshop on Learning with Multiple Views}, Bonn, Germany.

\bibitem[{Brown et~al.(2020)Brown, Mann, Ryder, Subbiah, Kaplan, Dhariwal, Neelakantan, Shyam, Sastry, Askell et~al.}]{gpt3}
Tom Brown, Benjamin Mann, Nick Ryder, Melanie Subbiah, Jared~D Kaplan, Prafulla Dhariwal, Arvind Neelakantan, Pranav Shyam, Girish Sastry, Amanda Askell, et~al. 2020.
\newblock Language models are few-shot learners.
\newblock \emph{Advances in neural information processing systems}, 33:1877--1901.

\bibitem[{Chung et~al.(2022)Chung, Hou, Longpre, Zoph, Tay, Fedus, Li, Wang, Dehghani, Brahma, Webson, Gu, Dai, Suzgun, Chen, Chowdhery, Castro-Ros, Pellat, Robinson, Valter, Narang, Mishra, Yu, Zhao, Huang, Dai, Yu, Petrov, Chi, Dean, Devlin, Roberts, Zhou, Le, and Wei}]{flan-t5}
Hyung~Won Chung, Le~Hou, Shayne Longpre, Barret Zoph, Yi~Tay, William Fedus, Yunxuan Li, Xuezhi Wang, Mostafa Dehghani, Siddhartha Brahma, Albert Webson, Shixiang~Shane Gu, Zhuyun Dai, Mirac Suzgun, Xinyun Chen, Aakanksha Chowdhery, Alex Castro-Ros, Marie Pellat, Kevin Robinson, Dasha Valter, Sharan Narang, Gaurav Mishra, Adams Yu, Vincent Zhao, Yanping Huang, Andrew Dai, Hongkun Yu, Slav Petrov, Ed~H. Chi, Jeff Dean, Jacob Devlin, Adam Roberts, Denny Zhou, Quoc~V. Le, and Jason Wei. 2022.
\newblock \href {http://arxiv.org/abs/2210.11416} {{Scaling Instruction-Finetuned Language Models}}.

\bibitem[{Derriere et~al.(2010)Derriere, Preite-Martinez, Richard, Cambrésy, and Padovani}]{ivoa}
Sébastien Derriere, Andrea Preite-Martinez, Alexandre Richard, Laurent Cambrésy, and Paolo Padovani. 2010.
\newblock \href {https://doi.org/10.5479/ADS/bib/2010ivoa.rept.0303D} {{Ontology of Astronomical Object Types Version 1.3}}.
\newblock International Virtual Observatory Alliance.

\bibitem[{Ding et~al.(2023)Ding, Qin, Liu, Chia, Li, Joty, and Bing}]{ding-etal-2023-gpt}
Bosheng Ding, Chengwei Qin, Linlin Liu, Yew~Ken Chia, Boyang Li, Shafiq Joty, and Lidong Bing. 2023.
\newblock \href {https://doi.org/10.18653/v1/2023.acl-long.626} {{Is {GPT}-3 a Good Data Annotator?}}
\newblock In \emph{Proceedings of the 61st Annual Meeting of the Association for Computational Linguistics (Volume 1: Long Papers)}, pages 11173--11195, Toronto, Canada. Association for Computational Linguistics.

\bibitem[{D'Souza and Auer(2022)}]{cs-ner}
Jennifer D'Souza and S{\"o}ren Auer. 2022.
\newblock \href {https://doi.org/https://doi.org/10.1007/978-3-031-21756-2_3} {{Computer Science Named Entity Recognition in the Open Research Knowledge Graph}}.
\newblock In \emph{From Born-Physical to Born-Virtual: Augmenting Intelligence in Digital Libraries}, pages 35--45, Hanoi, Vietnam. Springer International Publishing.

\bibitem[{D’Souza(2023)}]{agri-ner}
Jennifer D’Souza. 2023.
\newblock \href {https://doi.org/10.20944/preprints202305.1393.v1} {{Agriculture Named Entity Recognition - Towards FAIR, Reusable Scholarly Contributions in Agriculture}}.
\newblock \emph{Preprint}.

\bibitem[{Enkhsaikhan et~al.(2021)Enkhsaikhan, Liu, Holden, and Duuring}]{geo-ner}
Majigsuren Enkhsaikhan, Wei Liu, Eun-Jung Holden, and Paul Duuring. 2021.
\newblock \href {https://doi.org/10.1007/s10115-020-01532-6} {{Auto-Labelling Entities in Low-Resource Text: A Geological Case Study}}.
\newblock \emph{Knowledge and Information Systems}, 63(3):695–715.

\bibitem[{Ghosh et~al.(2022)Ghosh, Santra, Iqbal, and Basuchowdhuri}]{ghosh-etal-2022-astro}
Madhusudan Ghosh, Payel Santra, Sk~Asif Iqbal, and Partha Basuchowdhuri. 2022.
\newblock \href {https://aclanthology.org/2022.wiesp-1.12} {{Astro-m{T}5: Entity Extraction from Astrophysics Literature using m{T}5 Language Model}}.
\newblock In \emph{Proceedings of the first Workshop on Information Extraction from Scientific Publications}, pages 100--104, Online. Association for Computational Linguistics.

\bibitem[{Grezes et~al.(2022)Grezes, Blanco-Cuaresma, Allen, and Ghosal}]{grezes-etal-2022-overview}
Felix Grezes, Sergi Blanco-Cuaresma, Thomas Allen, and Tirthankar Ghosal. 2022.
\newblock \href {https://aclanthology.org/2022.wiesp-1.1} {{Overview of the First Shared Task on Detecting Entities in the Astrophysics Literature ({DEAL})}}.
\newblock In \emph{Proceedings of the first Workshop on Information Extraction from Scientific Publications}, pages 1--7, Online. Association for Computational Linguistics.

\bibitem[{Hedderich et~al.(2021)Hedderich, Lange, Adel, Str{\"o}tgen, and Klakow}]{hedderich-etal-2021-survey}
Michael~A. Hedderich, Lukas Lange, Heike Adel, Jannik Str{\"o}tgen, and Dietrich Klakow. 2021.
\newblock \href {https://doi.org/10.18653/v1/2021.naacl-main.201} {A survey on recent approaches for natural language processing in low-resource scenarios}.
\newblock In \emph{Proceedings of the 2021 Conference of the North American Chapter of the Association for Computational Linguistics: Human Language Technologies}, pages 2545--2568, Online. Association for Computational Linguistics.

\bibitem[{Longpre et~al.(2023)Longpre, Hou, Vu, Webson, Chung, Tay, Zhou, Le, Zoph, Wei, and Roberts}]{pmlr-v202-longpre23a}
Shayne Longpre, Le~Hou, Tu~Vu, Albert Webson, Hyung~Won Chung, Yi~Tay, Denny Zhou, Quoc~V Le, Barret Zoph, Jason Wei, and Adam Roberts. 2023.
\newblock \href {https://proceedings.mlr.press/v202/longpre23a.html} {{The Flan Collection: Designing Data and Methods for Effective Instruction Tuning}}.
\newblock In \emph{Proceedings of the 40th International Conference on Machine Learning}, volume 202 of \emph{Proceedings of Machine Learning Research}, pages 22631--22648. PMLR.

\bibitem[{Murphy et~al.(2006)Murphy, McIntosh, and Curran}]{murphy-etal-2006-named}
Tara Murphy, Tara McIntosh, and James~R. Curran. 2006.
\newblock \href {https://aclanthology.org/U06-1010} {Named entity recognition for astronomy literature}.
\newblock In \emph{Proceedings of the Australasian Language Technology Workshop 2006}, pages 59--66, Sydney, Australia.

\bibitem[{Wang et~al.(2021)Wang, Liu, Xu, Zhu, and Zeng}]{wang-etal-2021-want-reduce}
Shuohang Wang, Yang Liu, Yichong Xu, Chenguang Zhu, and Michael Zeng. 2021.
\newblock \href {https://doi.org/10.18653/v1/2021.findings-emnlp.354} {Want to reduce labeling cost? {GPT}-3 can help}.
\newblock In \emph{Findings of the Association for Computational Linguistics: EMNLP 2021}, pages 4195--4205, Punta Cana, Dominican Republic. Association for Computational Linguistics.

\bibitem[{Xue et~al.(2021)Xue, Constant, Roberts, Kale, Al-Rfou, Siddhant, Barua, and Raffel}]{mt5}
Linting Xue, Noah Constant, Adam Roberts, Mihir Kale, Rami Al-Rfou, Aditya Siddhant, Aditya Barua, and Colin Raffel. 2021.
\newblock \href {https://doi.org/10.18653/v1/2021.naacl-main.41} {{m{T}5: A Massively Multilingual Pre-trained Text-to-Text Transformer}}.
\newblock In \emph{Proceedings of the 2021 Conference of the North American Chapter of the Association for Computational Linguistics: Human Language Technologies}, pages 483--498, Online. Association for Computational Linguistics.

\end{thebibliography}
\bibliographystyle{acl_natbib}

\end{document}